\documentclass{article}
\usepackage{times}
\usepackage{hyperref}
\usepackage{url}
\usepackage[english]{babel}
\usepackage[utf8]{inputenc}
\usepackage{csquotes}
\usepackage{amsmath} 
\usepackage{amsthm,amssymb,latexsym,epic,eepic,epsfig,graphics,psfrag}
\usepackage{amsfonts}
\usepackage{bm}
\usepackage{float}
\usepackage{booktabs}
\usepackage{array,multirow}
\usepackage{xspace}

\newcommand{\unsupfull}{\texttt{UL}\xspace}
\newcommand{\unsupsmall}{\texttt{UL}$_{\rm small}$\xspace}
\newcommand{\supfull}{\texttt{SL}\xspace}
\newcommand{\supsmall}{\texttt{SL}$_{\rm small}$\xspace}
\newcommand{\msmall}{$M_{\rm small}$\xspace}
\newcommand{\nsmall}{$N_{\rm small}$\xspace}
\newcommand{\ntrain}{\ensuremath{N_{\rm train}}}
\newcommand{\ntest}{\ensuremath{N_{\rm test}}}

\title{Dimensionality reduction for click-through rate prediction: Dense versus sparse representation}
\author{
	Bjarne {\O}rum Fruergaard, Toke Jansen Hansen\\
	Adform ApS\\
	Hovedvagtsgade 6\\
	DK-1103 Copenhagen K, Denmark\\
	\texttt{\{b.fruergaard,t.hansen\}@adform.com}\\
	\and
	Lars Kai Hansen\\
	Technical University of Denmark\\
	Anker Engelunds Vej 1,\\
	DK-2800 Kgs. Lyngby, Denmark\\
	\texttt{lkai@dtu.dk}\\
	}
\begin{document}

\maketitle

\begin{abstract}
	In online advertising, display ads are increasingly being placed based on real-time auctions where the advertiser who wins gets to serve the ad. This is called real-time bidding (RTB). In RTB, auctions have very tight time constraints on the order of 100ms. Therefore mechanisms for bidding intelligently such as click-through rate prediction need to be sufficiently fast. In this work, we propose to use dimensionality reduction of the user-website interaction graph in order to produce simplified features of users and websites that can be used as predictors of click-through rate. We demonstrate that the Infinite Relational Model (IRM) as a dimensionality reduction offers comparable predictive performance to conventional dimensionality reduction schemes, while achieving the most economical usage of features and fastest computations at run-time. For applications such as real-time bidding, where fast database I/O and few computations are key to success, we thus recommend using IRM based features as predictors to exploit the recommender effects from bipartite graphs.
\end{abstract}

\newpage

\section{Introduction}
In advertising, one is interested in segmenting people and targeting ads based on segments \cite{Apte2001}. With the rapid growth of the Web as a publishing platform, new advertising technologies have evolved, offering greater reach and new possibilities for targeted advertising. One such innovation is real-time bidding (RTB), where upon a user's request for a specific URL, an online real-time auction is started amongst numerous participants, competing to serve their advertisement. The participants are allotted a limited time on the order of 100ms to query their data sources and come up with a bid, and the winner gets to display their advertisement. Thus if the computational complexity can be reduced, more complex decision processes can be invoked. In this work, we evaluate how dimensionality reduction can be used to simplify predictors of click-through rate.

We focus on three techniques for dimensionality reduction of the large bipartite graph of user-website interactions, namely Singular Value Decomposition (SVD) \cite{Golub1965}, Non-negative Matrix Factorization (NMF) \cite{Lee1999}, and the Infinite Relational Model (IRM) \cite{Kemp2006}. We are interested in how the different levels of sparsity of the output features imposed by each of the models affect the performance in a click-through rate prediction task. In the RTB setup, where low latency and high throughput are both of crucial importance, database queries need to require as little I/O as possible, and computing model predictions need to involve as few operations as possible. Therefore a good idea is to ``compress'' very high-cardinality features using dimensionality reduction techniques and at the same time potentially benefit from recommender effects \cite{Hillard2011}. This presents a trade-off between how much to compress in order to speed up I/O and calculations versus retaining, or exceeding, the performance of a high cardinality feature.

By investigating the SVD, NMF, and the IRM, we essentially vary the compression of a high-cardinality feature (user-website engagements). The SVD produces dense singular vectors, thus requiring the most I/O as well as computation. The NMF is known to produce sparse components \cite{Lee1999}, meaning that zeros need not be stored, retrieved nor used in computations, and thus requires less I/O and computation. The IRM offers the most sparse representation, in that it produces hard cluster assignments, hence I/O and computation are reduced to a single weight per mode.


We present results that use either of the dimensionality reduction techniques' outputs as predictors for a click-through rate prediction task. Our experiments show that a compact representation based on the NMF outperforms the other two options. If one however wants to use as little I/O and as simple computations as possible, the very compact representation from the IRM model offers an interesting alternative. While incurring a limited loss of lift relative to the NMF, the IRM based predictors yield the fastest training speed of the downstream logistic regression classifier and also results in the most economical usage of features and fastest possible computations at run-time. The IRM further has the advantage that it alleviates the need for model order selection, which is required in NMF. While the dense features produced by SVD also find usage in terms of predictive performance, the dense features inhibit the logistic regression training time, and if low database I/O as well as fast computation of predictions is a priority, the SVD will not be of great use.

A key enabling factor in running the IRM with the data we present in this work, is a sampler written for the graphics processing unit (GPU) \cite{Hansen2011a}, without which learning of the IRM model would not be feasible, at least not on a day-by-day schedule. To demonstrate the feasibility of the IRM as a large-scale sparse dimensionality reduction, we run final tests on a full-scale click-through rate  data set and compare the performances with not using any dimensionality reductions.

\subsection{Related work}
Within the area of online advertising, computational targeting techniques are often faced with the challenge of very few observations per feature, particularly of the positive label (i.e., click, action, buy). A common approach to alleviate such label sparsity is to use collaborative filtering type algorithms, where one allow similar objects to ``borrow'' training data and thus constrain the related objects to have similar predicted behaviour. Studies hereof are common for sponsored search advertising where the objects of interest are query-ad pairs \cite{Hillard2011,Dave2010a}, but the problem is similar to that of user-website pairs that we study. To our knowledge we are the first to report on the usage of the IRM co-clustering of user-website pairs and the results should be applicable for query-add click-through rate prediction as well.

By representing users in a compressed or latent space based on the user-website graph, we are essentially building profiles of users based on their behaviour and using those profiles for targeted advertising. This approach is well studied with many other types of profiles based on various types of information: For using explicit features available for predicting click-through rates, \cite{Richardson2007a} is a good resource: Latent factor models have been proposed to model click-through rates in online advertising, see e.g. \cite{Menon2011a}: For examples of using dimensionality reduction techniques in the construction of click-through rate models, such as the NMF, see \cite{Chen2009}. We believe our contribution to have applications in many such setups, either as an additional predictor or for incorporation as \textit{a priori} information (priors, constraints, etc.) which can help with identifiability of the models.

We regard the problem of predicting click-through rates as a supervised learning task, i.e., given historical observations with features (or predictors) available about the user, webpage, and ad, along with the labels of actions (in our case click (1) or not-click (0)), the task is to learn a classifier for predicting unseen observations, given the features. This is the approach taken also by e.g., \cite{Richardson2007a}. As in \cite{Richardson2007a}, we build a probabilistic model based on logistic regression for predicting click-through rates. What we add, is additional features based on dimensionality reduction, as well as a sparsity inducing constraint based on the $L_1$-norm.

\section{Methods}\label{sec:Method}

We are interested in estimation of features which can improve click-through rate predictions. In this work, we focus on introducing features from different dimensionality reduction techniques based on a bipartite graph of users and websites (URLs), and using them in a simple probabilistic model for click-through rate prediction, namely logistic regression. In the following, we introduce the dimensionality reduction techniques which we evaluate.

\subsection{Dimensionality reduction techniques}

\subsubsection{Singular value decomposition}\label{sec:SVD}
The singular value decomposition (SVD) of a rank $R$ matrix $\bm X$ is given as the factorization
$\bm X = \bm U \bm \Sigma \bm V^\top = \sum_{i=1}^R \sigma_i \bm u_i \bm v_i^\top$,
where $\bm U$ and $\bm V$ are unitary matrices $\bm U^\top \bm U = \bm V^\top \bm V = \bm I$ and hold the left and right singular vectors of $\bm X$, respectively. The diagonal matrix $\bm \Sigma$ contains the singular values of $\bm X$. By selecting only the $K$ largest singular values of $\bm \Sigma$, i.e., truncating all other singular values to zero, one obtains the approximation
$\bm{\tilde X} = \bm U \bm{\tilde \Sigma} \bm V^\top = \sum_{i=1}^K \sigma_i \bm u_i \bm v_i^\top$,
which is the rank $K$ optimal solution to $\text{arg}\min ||\bm X-\bm{\tilde X}||_2^2$. This truncation corresponds to disregarding the $R-K$ dimensions with the least variances of the bases $\bm U$ and $\bm V^\top$ as noise.

\subsubsection{Non-negative matrix factorization}\label{sec:NMF}
Non-negative matrix factorization (NMF) received its name as well as its popularity in \cite{Lee1999}. NMF is a matrix factorization comparable to SVD, the crucial difference being that NMF decomposes into non-negative factors and impose no orthogonality constraints. Given a non-negative input matrix $\bm X$ with dimensions $M \times N$, NMF approximates the decomposition $\bm X \approx \bm W \bm H$, where $\bm W$ is an $M \times K$ non-negative matrix, $\bm H$ a $K \times N$ non-negative matrix, and $K$ is the number of components. By selecting $K << min(M,N)$ one approximates the decomposition of $\bm X^{(M \times N)} = \bm W^{(M \times K)} \bm H^{(K \times N)} + \bm E^{(M \times N)}$, thereby disregarding some residual (unconstrained) matrix $\bm E$ as noise.

NMF has achieved good empirical results as an unsupervised learning technique within many applications, e.g., for document clustering \cite{Berry2006,Xu2003,Wahlgreen2011}, visual coding \cite{Lee1999}, and bioinformatics \cite{Dong2005}. For NMF applications for computational advertising, see also \cite{Chen2009}.

\subsubsection{Infinite relational model}\label{sec:IRM}
The Infinite Relational Model (IRM)
has been proposed as a Bayesian generative model for graphs. Generative models can provide accurate predictions and through inference
of relevant latent variables they can inform the user about mesoscale structure.
The IRM model can be cast as co-clustering approach for bipartite networks
where the nodes of each mode are grouped simultaneously. A benefit of the IRM model over
existing co-clustering approaches is that the model explicitly exploit the statistical properties
of binary graphs and allows the number of components of each mode to be inferred from the data.

The generative process for the Relational Model
\cite{Kemp2006,Xu2006,Nowicki2001} is given by:\\
$\diamond$ Sample the row cluster probabilities, i.e., $\bm \mu^{(1)}\sim Dirichlet(\alpha^{(1)}/K^{(1)}\bm e^{(1)})$.\\
$\diamond$ Sample row cluster assignments, i.e., $m=1,\ldots, M$ $\bm z_m^{(1)}\sim Discrete(\bm \mu^{(1)})$. \\
$\diamond$ Sample the column cluster probabilities, i.e., $\bm \mu^{(2)}\sim Dirichlet(\alpha^{(2)}/K^{(2)}\bm e^{(2)})$.\\
$\diamond$ Sample column cluster assignments, i.e., $n=1,\ldots, N$ $\bm z_n^{(2)}\sim
Discrete(\bm\mu^{(2)})$.\\
$\diamond$ Sample between cluster relations, i.e., $i=1,\ldots, I$ and $j=1,\ldots, J$ $\eta_{ij}\sim Beta(\beta^+,\beta^-)$.\\
$\diamond$ Generate links, i.e., $m=1,\ldots, M$ and $n=1,\ldots, N$  $X_{nm}\sim
Bernoulli(\bm z_n^{(1)\top}\bm\eta\bm z_m^{(2)})$.\\
Where $K^{(1)}$ and $K^{(2)}$ denote the number of row and column clusters respectively whereas $\bm e^{(1)}$ and $\bm e^{(2)}$ are vectors of ones with size $K^{(1)}$ and $K^{(2)}$.
The limits $K^{(1)}\rightarrow \infty$ and $K^{(2)}\rightarrow \infty$ lead to the Infinite Relational Model (IRM) which has an analytic solution given by the Chinese
Restaurant Process (CRP) \cite{Xu2006,Kemp2006,Neal2000}.

Rather than collapsing the parameters of the model, we apply blocked sampling that allows for parallel GPU computation \cite{Hansen2011a}. Moreover, the CRP is approximated by the truncated stick breaking construction (TSB), and the truncation error becomes insignificant when the model is estimated for large values of $K^{(1)}$ and $K^{(2)}$, see also \cite{Xu2007}.

\subsection{Supervised learning using logistic regression}\label{sec:LR}

For learning a model capable of predicting click-through rates trained on historical data, we employ logistic regression with sparsity constraints; for further details see for instance \cite{Bishop2007a, Andrew2007}. Given data consisting of $n=1,\ldots,N$ observations with $p$-dimensional feature vectors $\bm x_n^\top$ and binary labels $y_n \in {0,1}$, the probability of a positive event can be modeled with the \textit{logistic function} and a single weight $\omega$ per feature. I.e., $p(Y_n=1 | \bm x_n, \bm \omega) = \sigma(\bm x_n^\top \bm \omega) = 1 / (1+\exp(-\bm x_n^\top \bm \omega))$, referred to as $p_n$ in the following. The optimization problem for learning the weights $\bm \omega$ becomes
\begin{align}
	\min_{\bm \omega} \quad \Omega_{L_1}(\bm \omega) - \sum_{n=1}^N y_d \log( p_n ) + (1 - y_d) \log(1 - p_n),\label{eq:logreg}
\end{align}
where $\Omega_{L_1} = \bm \lambda^\top |\bm \omega|_1 = \sum_{i=1}^p \lambda_i |\omega_i|$ is added to control overfitting and produce sparse solutions. For skewed target distributions, an intercept term $\omega_0$ may be included in the model by appending an all-one feature to all observations. The corresponding regularization term $\lambda_0$ then needs to be fixed to zero.

For training the logistic regression model, one can use gradient-descent type optimizers and quasi-Newton based algorithms are a popular choice. With $L_1$-penalty, however, a little care must be taken since off-the-shelf Newton-based solvers require the objective function to be differentiable, which \eqref{eq:logreg} is not due to the penalty function which is not differentiable in zero. In this work we base our logistic regression training on OWL-QN \cite{Andrew2007} for batch learning. For online learning using stochastic gradient descent with $L_1$-penalization, see \cite{Tsuruoka2009}.

Performing predictions with a logistic regression model is as simple as computing the logistic function on the features of a test observation, $\bm{\tilde x}$. In terms of speed, however, it matters how the features of $\bm{\tilde x}$ are represented. In particular for a binary feature vector $\bm x$
\begin{align}
	\sigma(\bm x^\top \bm \omega) = \frac{\exp(\bm x^\top \bm \omega)}{1 + \exp(\bm x^\top \bm \omega)} = \frac{\prod_{i=1}^p \exp(x_i\omega_i)}{1 + \prod_{i=1}^p \exp(x_i\omega_i)} &\stackrel{\text{$\bm x$ binary}}{=} \frac{\prod_{i':x_{i'}=1} \exp(\omega_{i'}) }{1 + \prod_{i':x_{i'}=1} \exp(\omega_{i'})} \label{eq:fast_lr_pred}
\end{align}
I.e., predicting for binary feature vectors scales in the number of non-zero elements of the feature vector, which makes computations considerably faster. Additionally, using the right-hand side of \eqref{eq:fast_lr_pred}, $\exp(\cdot)$ can be performed when storing the weights in memory or a database, hences saves further processing power. This has two consequences: 1) Binary features are more desirable for making real-time predictions and 2) the sparser the features, the less computation time and I/O from databases is required.

\section{Experiments}

The data we use for our experiments originate from Adform's ad transaction logs. In each transaction, e.g., when an ad is served, the URL where the ad is being displayed and a unique identifier of the users web browser is stored along with an identifier of the ad. Likewise, a transaction is logged when a user clicks an ad. From these logs, we prepare a data set over a period of time and use the final day for testing and use the rest for training.

As a pre-processing step, all URLs in the transaction log are stripped of any query-string that might be trailing the URL\footnote{Query-string: Anything trailing an ``?'' in a URL, including the ``?''.}, however the log data are otherwise unprocessed.

\subsection{Dimensionality reduction}\label{sec:dimens_reduct}
From the training set transactions, we produce a binary bipartite graph of users in the first mode and URLs in the second mode. This is an unweighted, undirected graph where edges represent which URLs a user has seen, i.e., we do not use the number of times the user has engaged each URL. The graph we obtain has $M$=9,304,402 unique users and $N$=7,056,152 unique URLs. We denote this graph \unsupfull.

As we will be repeating numerous supervised learning experiments, that each can be quite time consuming for the entire training set, we do our main analysis based on experiments from a subset of transactions. As an inclusion criteria, we select the top \msmall=99,854 users based on the number of URLs they have seen and URLs with visits from at least 100 unique users, resulting in \nsmall=70,436 URLs being included. Based on those subsets of users and URLs, we produce a smaller transaction log, from which we also construct a bipartite graph denoted \unsupsmall.

\subsubsection{Method details}
For the sampled data for unsupervised learning, \unsupsmall, we use the different dimensionality reduction techniques presented in Section \ref{sec:Method} to obtain new per-user and per-URL features.

For obtaining the SVD-based dense left and right singular vectors, we use \texttt{SVDS} included with Matlab to compute the 500 largest eigenvalues with their corresponding eigenvectors. In the supervised learning, by joining our data by user and URL with the left and right singular vectors, respectively, we can use anything from 1 to 500 of the largest eigenvectors for each modality as features.

We use the NMF Matlab Toolbox from \cite{Li2013} to decompose \unsupsmall into non-negative factors. We use the original algorithm introduced in \cite{Lee1999} with the least-squares objective and multiplicative updates (\texttt{nmfrule} option in the NMF Toolbox). With NMF we need to decide the model order, i.e., number of components to fit in each of the non-negative factors. Hence, to investigate the influence of NMF model order, we train NMF using various model orders of 100, 300, and 500 number of components. We run the toolbox with the default configurations for convergence tolerance and maximum number of iterations.

As detailed in Section \ref{sec:IRM}, we use the GPU sampling scheme from \cite{Hansen2011a} for massively speeding up the computation of the IRM model. The IRM estimation infers the number of components (i.e., clusters) separately for each modality, however, it does require we input a maximum number of components for users and URLs. For \unsupsmall, we run with $K_{\rm max}$=500 for both modalities and terminate the estimation after 500 iterations. The IRM infers 216 user clusters and 175 URL cluster for \unsupsmall, i.e., well below the $K_{\rm max}$ we specify. 

For the full dataset \unsupfull, we have only completed the dimensionality reduction using IRM, which is thanks to our access to the aforementioned GPU sampling code. Again we run the IRM for 500 iterations, and with 500 as $K_{\rm max}$ for each modality. The IRM infers 408 user clusters and 380 URL clusters for \unsupfull; again well below $K_{\rm max}$. 

Running the SVD and NMF for a data set the size of \unsupfull within acceptable times (i.e., within a day or less), is in it self a challenge and requires specialized software, either utilizing GPUs or distributed computation (or both). As we have not had immediate access to any implementations capable hereof, the SVD and NMF decompositions of \unsupfull remain as future work. Hence, for click-through rate prediction on the full data set, we demonstrate only the benefit of using the IRM cluster features over not using any dimensionality reduction.

\subsection{Supervised learning}
For testing the various dimensionality reductions, we construct several training and testing data sets from RTB logs with observations labeled as click (1) or non-click (0). The features we use are summarized in table \ref{table:predictors}.

\begin{table}[!t]
	\small
	\centering
	\caption{Names and descriptions of the predictors used to predict click-through rates.}
	\label{table:predictors}
	\begin{tabular}{llp{.63\textwidth}}
		\\
		Ref & Feature(s) & Description \\
		\toprule
		$f_1$ & (BannerId, Url) & A one-of-K encoding of the cross-features between $BannerId$ and $Url$, which indicates where a request has been made. This serves as a baseline predictor in all of our experiments. \\
		$f_2$ & UrlsVisited & A vector representation (zeros and ones) of URLs that a specific user has visited in the past. \\ 
		$f_3$ & UserCluster & A one-of-K encoding of which IRM cluster a specific user belongs to. \\ 
		$f_4$ & UrlCluster & A one-of-K encoding of which IRM cluster a specific URL belongs to. \\
		$f_5^{(n)}$ & UserSVD$n$Loading & The continous-valued $n$-dimension left singular vector of a specific user from the SVD. \\ 
		$f_6^{(n)}$ & UrlSVD$n$Loading & The continous-valued $n$-dimension right singular vector of a specific URL from the SVD. \\ 
		$f_7^{(n)}$ & UserNMF$n$Loading & The continous-valued cluster assignment vector of a specific user according to the NMF-$n$ decomposition. \\ 
		$f_8^{(n)}$ & UrlNMF$n$Loading & The continous-valued cluster assignment vector of a specific URL according to the NMF-$n$ decomposition. \\
		\bottomrule 
	\end{tabular}
\end{table}

Based on the full set of users and URLs as well as the sub-sampled sets, detailed in Section \ref{sec:dimens_reduct}, we prepare training and testing data sets based on the features of Table \ref{table:predictors} for our logistic regression classifier. We denote the full dataset \supfull and the sampled \supsmall. The data are represented as $N \times p$ matrices, i.e., with columns being features and rows being observations.

\subsubsection{Method details}

From the predictors of Table \ref{table:predictors}, we train a number of logistic regression classifiers, using $L_1$-penalization for sparsity, see also Section \ref{sec:LR}. For the stopping criteria, we run until the change of the objective value between iterations falls below 1e-6. As the classes (clicks vs. non-clicks) are highly unbalanced, we also learn an unpenalized intercept term. In order not to introduce any advantages (or disadvantages) to some predictors over others, we do not normalize the input features for any of the predictors in any way. Rather, we first select one regularization strength, $\lambda_{f_1}$, for the baseline predictor only, $f_1$, and fix that through all other trials. In each experiment, we then use other predictors $f_3$-$f_8$ in addition to $f_1$ and select another regularization strength, $\lambda_{f_{\geq 3}}$, jointly regularizing those predictors, but with $\lambda_{f_1}$ still fixed for $f_1$. We compare to using $f_2$ regularized by $\lambda_{f_2}$ in addition to $f_1$ and henceforth refer to this model as NODR, short for no dimensionality reduction.

For each trained model, we measure the performance in terms of the negative Bernoulli log-likelihood (LL), which measures the mismatch between the observations and the predictions of the model, i.e., the lower, the better. The likelihoods we report are normalized with respect to the baseline likelihood of the click-through rate evaluated on the test set, such that in order to outperform the baseline, they should fall between 0 and 1.

\begin{table}[!b]
	\small
	\centering
	\caption{Statistics of the various predictors on the sampled data set.}
	\label{table:pred_usage}
	\begin{tabular}{lccc}
		&&&\\
		Feature & $p$ & nnz & sparsity  \\ 
		\toprule
		$f_1$ & 44086 & 143120 & 1 - 2.3e-5 \\ 
		$f_2$ & 42910 & 8824491 & 1 - 1.4e-3 \\ 
		$f_3$ & 216 & 143120 & 1 - 4.6e-3 \\
		$f_4$ & 175 & 143120 & 1 - 5.7e-3 \\
		$f_5$,$f_6$ & 100 / 300 / 500 & $dense$ & 0  \\
		$f_7$ & 100 / 300 / 500 & 4745568 / 9780078 / 13993847 & 0.67 / 0.77 / 0.80 \\
		$f_8$ & 100 / 300 / 500 & 4174552 / 14363612 / 23712222 & 0.71 / 0.67 / 0.67\\
		\bottomrule
	\end{tabular} 		
\end{table}

\subsection{Results on \supsmall}

\begin{table}[!t]
	\footnotesize
	\centering
	\caption{Results for the sub-sampled data set.}
	\label{table:results1}
	\begin{tabular}{cl@{\hspace{1em}(}c@{,\hspace{0.4em}}c@{,\hspace{0.4em}}c@{)\hspace{1em}}ccccc@{}l}
		\\
		& Model & $\lambda_{f_1}$ & $\lambda_{f_2}$ & $\lambda_{f_{\geq3}}$ & Time (s) & nnz$_{all}$ & nnz$_{f_2}$ & LL$\cdot 100$ & \% Lift \\
		\toprule
		& $f_1$ & 0.8 & - & - & 9 & 3612 & - & 93.83 & 0.00 \\ 
		\midrule
		\parbox[t]{2mm}{\multirow{3}{*}{\rotatebox[origin=c]{90}{NODR}}}\\
		& $f_1$, $f_2$ & 0.8 & 10.6 & - & 91 & 3943 & 760 & 88.15 & 6.05 \\
		\\
		\midrule
		\parbox[t]{2mm}{\multirow{3}{*}{\rotatebox[origin=c]{90}{IRM}}} & $f_1$, $f_3$ & 0.8 & - & 6.0e-4 & 13 & 3653 & - & 90.19 & 3.88 \\
		&$f_1$, $f_3,f_4$ & 0.8 & - & 7.0e-4 & 16 & 3674 & - & \textbf{89.84} & \textbf{4.25} &  $\bm\triangledown$ \\ 
		\cmidrule{2-11}
		& $\bm\triangledown+f_2$ & 0.8 & 15.4 & 7.0e-4 & 76 & 3861 & 366 & 87.78 & 6.45\\
		\midrule
		\parbox[t]{2mm}{\multirow{7}{*}{\rotatebox[origin=c]{90}{SVD}}}&$f_1$, $f_5^{(100)}$ & 0.8 & - & 0.1 & 19 & 3479 & - & 89.87 & 4.22\\ 
		&$f_1$, $f_5^{(300)}$ &  0.8 & - & 0.3 & 29 & 3502 & - & 89.73 & 4.37 \\ 
		&$f_1$, $f_5^{(500)}$ &  0.8 & - & 0.3 & 56 & 3552 & - & 89.73 & 4.37 \\
		&$f_1$, $f_5^{(100)}$, $f_6^{(100)}$ & 0.8 & - & 7.0e-4 & 649 & 3409 & - & 89.15 & 4.99 \\ 
		&$f_1$, $f_5^{(300)}$, $f_6^{(300)}$ & 0.8 & - & 7.0e-4 & 2487 & 3702 & - & \textbf{88.92} & \textbf{5.23} & $\bm\diamond$ \\ 
		&$f_1$, $f_5^{(500)}$, $f_6^{(500)}$ & 0.8 & - & 1.2e-3 & 4082 & 4027 & - & 89.55 & 4.56 \\ 
		\cmidrule{2-11}
		& $\bm\diamond+f_2$ & 0.8 & 10.8 & 7.0e-4 & 3291 & 4063 & 484 & 87.90 & 6.32	\\
		\midrule
		\parbox[t]{2mm}{\multirow{7}{*}{\rotatebox[origin=c]{90}{NMF}}}&$f_1$, $f_7^{(100)}$ & 0.8 & - & 6.0e-3 & 30 & 3453 & - & 89.38 & 4.74 \\ 
		&$f_1$, $f_7^{(300)}$ & 0.8 & - & 3.0e-3 & 40 & 3467 & - & 89.15 & 4.99 \\ 
		&$f_1$, $f_7^{(500)}$ & 0.8 & - & 2.0e-3 & 45 & 3521 & - & 88.68 & 5.49 \\
		&$f_1$, $f_7^{(100)}$, $f_8^{(100)}$ & 0.8 & - & 5.0e-3 & 151 & 3389 & - & 89.05 & 5.09 \\ 
		&$f_1$, $f_7^{(300)}$, $f_8^{(300)}$ & 0.8 & - & 6.0e-3 & 392 & 3468 & - & \textbf{87.89} & \textbf{6.33} & $\bm\circ$\\ 
		&$f_1$, $f_7^{(500)}$, $f_8^{(500)}$ & 0.8 & - & 4.0e-3 & 740 & 3635 & - & 93.59 & 0.26 \\
		\cmidrule{2-11}
		& $\bm\circ+f_2$ & 0.8 & 11.2 & 6.0e-3 & 641 & 3973 & 680 & 86.91 & 7.38 \\
		\bottomrule
	\end{tabular}
\end{table}

For the sampled data the number of observations are as follows: \ntrain=138,847 and \ntest=4,273. In order to give the reader an idea about the dimensionalities of the features as well as their sparsity, in Table \ref{table:pred_usage} we summarize some numbers on the predictors on the sampled data set. For features $f_1$,$f_3$, and $f_4$, the number of non-zeros (nnz) and sparsities are somewhat trivial, since these are categorical features represented as one-of-K binary vectors. For the SVD features, $f_5$ and $f_6$, we see that the feature vectors become completely dense. For the NMF features, however, we can confirm the methods' ability to produce sparse components, i.e., only between 20-33\% of the components turn up as non-zeros, yet they are far from the sparsities of the IRM cluster features, $f_3$ and $f_4$.

\begin{table}[!b]
	\small
	\centering
	\caption{Results for the full data set.}
	\label{table:results_full}
	\begin{tabular}{cl@{\hspace{1em}(}c@{,\hspace{0.4em}}c@{,\hspace{0.4em}}c@{)\hspace{1em}}ccccc@{}l}
		\\
		& Model & $\lambda_{f_1}$ & $\lambda_{f_2}$ & $\lambda_{f_{\geq3}}$ & Time (s) & nnz$_{all}$ & nnz$_{f_2}$ & LL$\cdot 100$ & \% Lift \\
		\toprule
		& $f_1$ & 0.7 & - & - & 34 & 14152 & - & 91.76 & 0.00 \\ 
		\midrule
		\parbox[t]{2mm}{\multirow{3}{*}{\rotatebox[origin=c]{90}{NODR}}}\\
		& $f_1$, $f_2$ & 0.7 & 10.2 & - & 195 & 15673 & 3010 & 88.71 & 3.32 \\
		\\
		\midrule
		\parbox[t]{2mm}{\multirow{2}{*}{\rotatebox[origin=c]{90}{IRM}}} & $f_1$, $f_3,f_4$ & 0.7 & - & 1.2e-3 & 51 & 13604 & - & 89.35 & 2.63 &  $\bm\triangledown$ \\ 
		\cmidrule{2-11}
		& $\bm\triangledown+f_2$ & 0.7 & 10.2 & 1.2e-3 & 293 & 16018 & 2939 & 88.19 & 3.89\\
		\bottomrule
	\end{tabular}
\end{table}

In Table \ref{table:results1}, we report the normalized likelihoods, lifts and test-set optimal regularization strengths $\lambda_{f_1}$ and $\lambda_{f_{\geq 3}}$, with varying features used for training. The lifts are all relative to model $f_1$. The penalization strength $\lambda_{f_1}=0.8$ is selected as the one maximizing the performance of the classifier using only $f_1$, and is kept fixed for all the other classifiers. Note, that generalization of the penalization terms is an issue we do not currently address. The time reported in the table are the seconds it takes to train the logistic regression classifier. nnz$_{all}$ and nnz$_{f_2}$ are the respective number of non-zero weights of the resulting classifier for all the features and the $f_2$ feature only


In order to be able to further elaborate on the pros and cons of using the various dimensionality reduction techniques as features in the logistic regression classifier, we carry out another set of experiments for the models highlighted (bold and marked $\bm{\triangledown},\bm{\diamond},\bm{\circ}$) in Table \ref{table:results1}. We fix the values of $\lambda_{f_1}$ and $\lambda_{f_{\geq 3}}$ to the values from $\bm{\triangledown},\bm{\diamond}$, and $\bm{\circ}$, respectively, and append $f_2$ as an additional feature with each model and then tune the regularization strength $\lambda_{f_2}$. The results are shown in the rows of Table \ref{table:results1} with the symbols $\bm{\triangledown}+f_2,\bm{\diamond}+f_2$, and $\bm{\circ}+f_2$ under ``Model''.

The final experiment we run is with the full data set where we only evaluate the IRM based features and compare those to not using any dimensionality reduction. The number of observations for train and test are \ntrain=5,460,229 and \ntest=188,867. The selection of regularization terms we do as in the previous experiments. The results are reported in Table \ref{table:results_full}.


\section{Discussion}
From Table \ref{table:results1} we first concentrate on the best models from each dimensionality reduction, i.e., the results highlighted in bold. Comparing the lifts, we see that the NMF-300 features perform roughly one \%-point better than the SVD-300 features, which then in turn perform roughly another \%-point better than the IRM cluster features. Comparing to the classifier using just $f_1$ and $f_2$, i.e., no dimensionality reduction, we see that only the NMF-based classifier achieves slightly higher lift. Hence, using SVD or IRM based features as \textit{a replacement} for the $f_2$ feature would result in worse predictions. Seeing the number of non-zero weights dropping from 3943 using $f_2$ to 3468 using both NMF-300 features, indicates that the NMF offers a more economical representation which can replace $f_2$ while not sacrificing performance. The performance gain of NMF-300 we expect is achieved by the implicit data grouping effects of NMF, i.e., recommender effects.

In terms of training speed, we see that while the IRM based features fare worst in terms of lift, the fact that each mode is a categorical value represented in a one-of-K binary vector makes the input matrix very sparse, which speeds up the training of our classifier significantly and the model trains at least an order of magnitude faster than the other dimensionality reduction techniques and even significantly faster than training the NODR model. Hence, if fast training is a priority, either no dimensionality reduction should be used or the IRM based features can be used, but at the cost of slightly lower lift.

We now turn to the results for the models $\bm{\triangledown}+f_2,\bm{\diamond}+f_2$, and $\bm{\circ}+f_2$ in Table \ref{table:results1}. Here we investigate how the learning of weights for the high-cardinality feature $f_2$ is affected when combined with each of the optimal settings from the reduced dimension experiments. Again, observing the lifts, the NMF-300 based features combined with $f_2$ obtains the highest lift. However, the IRM based features now outperform the SVD ones and using either of the techniques in combination with $f_2$, we are able to obtain higher lifts than using only $f_2$.

For the training speed, we again see that the training using IRM features is by far the fastest amongst SVD and NMF and it is still faster than using $f_2$ only. What is more interesting, is the resulting number of non-zero weights, both in total and in the $f_2$ feature alone. Of all the different dimensionality reductions as well as NODR, using the IRM based representation requires the fewest non-zero weights at its optimal settings. Additionally, recalling from Section \ref{sec:LR}, that predictions can be made computationally very efficient, when the input features are binary indicator vectors, the IRM becomes all the more tractable. By combining the IRM based features with the explicit predictors $f_1$ and $f_2$, our classifier is able to improve the lift over not using dimensionality reduction while reducing the need for fetching many weights for predictions and with only a small reduction in lift, compared to the more computationally expensive classifiers based on NMF and SVD.

Finally, in Table \ref{table:results_full} we have run experiments using just the IRM based predictors with the full data set. The results confirm our findings from Table \ref{table:results1} and at the same time demonstrates both the feasibility of processing very large bipartite graphs using IRM as well as the application of the user and URL clusters as predictors of click-through rates.

\section{Conclusion}
We have presented results that demonstrate the use of three bimodal dimensionality reduction techniques, SVD, NMF, and IRM, and their applications as predictors in a click-through rate data set. We show that the compact representation based on the NMF is, in terms of predictive performance, the best option. For applications where fast predictions are required, however, we show that the binary representation from the IRM model is a viable alternative. The IRM based predictors yield the fastest training speed in the supervised learning stage, produces the most sparse model and offers the fastest computations at run-time, while incurring only a limited loss of lift relative to the NMF. In applications such as real-time bidding, where fast database I/O and few computations are key to success, we recommend using IRM based features as predictors.

\newpage
{\small
\subsubsection*{References}
\def\section*#1{}
\bibliography{NIPS_2013_Adform_IRM_paper,tjh}
}

\end{document}